%% file: reference.tex
\documentclass[11 pt]{llncs}
\pdfoutput=1
\usepackage[colorlinks, urlcolor=blue, citecolor=blue]{hyperref}
\usepackage{pifont}
\usepackage{amssymb}
\usepackage{amsmath}
\usepackage{subcaption}
\usepackage{kantlipsum} %<- For dummy text
\usepackage{mwe}
\usepackage{adjustbox}
\usepackage{wrapfig}
\usepackage{float}
\usepackage[utf8x]{inputenc}

\newcommand{\eg}{{\it e.g., }}
\newcommand{\etal}{{\it et~al. }}
\newcommand{\ie}{{\it i.e., }}
\usepackage{flexisym}
\usepackage{multirow}
\usepackage{mathtools, setspace,lscape,verbatim,color, alltt, amsfonts, wrapfig, boxedminipage,pifont,mathtools }
\usepackage{dblfloatfix}
\usepackage{aliascnt}
\usepackage{pifont}% http://ctan.org/pkg/pifont
\newcommand{\cmark}{\ding{51}}%
\newcommand{\xmark}{\ding{55}}%

\usepackage{xcolor}
\usepackage{balance}
\usepackage{fancyhdr, lastpage}
\newcommand{\name}{DeepFake}
\usepackage{lmodern}

\begin{document}

\title{DeepFakes: Detecting Forged and Synthetic Media Content Using Machine Learning}

\author{Sm Zobaed\inst{1} \and
Md Fazle Rabby\inst{1} \and
Md Istiaq Hossain\inst{2}\and
Ekram Hossain\inst{1}\and
Sazib Hasan\inst{3} \and
*Asif Karim\inst{4}\and
Khan Md. Hasib\inst{5}
}
\institute{University of Louisiana, Lafayette, LA 70503, USA \and
Southern Utah University, Cedar City, UT 84720, USA \and
Dixie State University, St. George, UT 84770, USA \and
Charles Darwin University, Casuarina, NT 0810, Australia \and
Ahsanullah University of Science \& Technology, Dhaka, Bangladesh
\\
\email{\{sm.zobaed1,ekram.hossain1\}@louisiana.edu},
\email{\{sourav.sust.cse.10, shishir2004x, khanmdhasib.aust,sazib25\}@gmail.com}, \email {asif.karim@cdu.edu.au} 
}

\authorrunning{Zobaed et al.}
%\tableofcontents
\titlerunning{\small{DeepFakes: Detecting Forged and Synthetic Media Content}}
\maketitle              % typeset the header of the contribution

\begin{abstract}
The rapid advancement in deep learning makes the differentiation of authentic and   
manipulated facial images and video clips unprecedentedly harder.
The underlying technology of manipulating facial appearances through deep generative approaches, enunciated as \textit{\name} that have emerged recently by promoting a vast number of malicious face manipulation applications. Subsequently, the need of other sort of techniques that can assess the integrity of digital visual content is indisputable to reduce the impact of the creations of~\name. A large body of research that are performed on \name~creation and detection create a scope of pushing each other beyond the current status. 
This study presents challenges, research trends, and directions related
to \name~creation and detection techniques by reviewing the notable research in the \name~domain to
facilitate the development of more robust approaches that could
deal with the more advance \name~in future.

\end{abstract}

\input{Sources/sec-int}

\input{Sources/sec-generation}

\input{Sources/sec-detection}
\input{Sources/sec-future}
\input{Sources/sec-concl}

%\linespread{.7}

\bibliographystyle{unsrt}
\bibliography{reference}
\end{document}

%% file: Sources/sec-int.tex
\section{Introduction}
\label{intro}

Because of the advances of deep learning and generative adversarial networks (GAN)~\cite{goodfellowgenerative}, creation of a realistically looking face image of a target person who really does not exist or altercation of facial appearance (attributes, identity, expression) is attainable with maintaining realism. The deep learning research community roughly refers to the technology as ``\name'' that is coined from ``deep learning'' and ``fake''. Generally, \name~approaches
require a massive volume of image and video data to train models
for generating realistic images and videos.
Because of the wide availability of robust pre-trained \name~models, malicious \name~contents are generated that create negative impact on the societies.     

The potential target of \name~is the public figures such as celebrities, priests, and politicians whose videos and images are largely available on the internet. More specifically, \name~is often used to alter faces of celebrities or politicians to other bodies in pornographic contents.
\name s can be abused to create political or ethnic tensions between countries to
fool common people to affect election, or create chaos in sports or global economy by creating fake contents.

There are numerous notable examples of \name~incidents that have been shared in the internet~\cite{fakeporn2,fakeporn,inci,kaliyar2020deepfake}. For instance, in 2018, a video posted in Facebook showing Former President of USA, Donald Trump taunted Belgium for remaining in the Paris climate agreement~\cite{inci}. By noticing the video clearly, it was determined that Trump’s hair looked stranger than usual and his voice was rolled up.
In 2019, a \name~video of Facebook owner, Mark Zuckerberg, was published on Instagram~\cite{inci}. In the video, Zuckerberg's speech was altered along with his facial expression so that the viewers can easily be distracted.
A recent release of an app named DeepNude raises issue since it is used to transform a person to a non-consensual pornography~\cite{guyfake}. 
Similarly, a Chinese app named ``Zao" got
viral lately for offering face swapping with
bodies of TV stars and even replace themselves into well-known
movies and TV clips~\cite{fakeappzao}. 
These forms of manipulation create a serious threat to privacy and identity, and even jeopardize personal lives.
Although the evil technology is undoubtedly a severe threat to world security, it is also used in positive purposes such as updating episodes of a visual content even after the actor is dead or creating speech of mute people. However, the number
of maliciously used cases \name~significantly outperforms the number of positively used cases.

The underlying mechanism for \name~creation is advanced
deep learning models such as autoencoders and GAN, which have been applied widely in the computer vision research community.
Due to the development of advanced deep networks
and the availability of a substantial amount of training data,
manipulated images and videos have turned out almost indistinguishable to human eyes
and even to robust algorithms. 
Hence, the creation of those manipulated contents becomes
simpler and takes comparatively lesser effort. This is because an identity image or small video clip of a targeted individual are sufficient for the inference tasks. 

The rise of stunning \name~creation vividly highlights the significance of judging the genuineness of digital media content. Because of the availability of various \name~creation tools, almost anyone can simply create forged content these days. As a result, in the computer vision research community, the study of \name~has gained traction in recent years for detecting such contents~\cite{lyu2020deepfake, guarnera2020preliminary,juefei2021countering,jafar2020forensics,trinh2020interpretable}. 
In~\cite{juefei2021countering}, Juefei-Xu \etal showed a distribution of \name~related papers in last 5 years, where $78\%$ of the total papers published in the last two years. This increase amount of paper in the last two years vividly highlights research interest revolved around \name s.

Governments and law enforcement are undertaking the spread of \name~creations with
new policies and regulations as well.
For example, US Senator named Ben Sasse proposed a bill titled \emph{S.3805 - Malicious deep fake prohibition act of 2018} in 2018 that 
introduces a new type of criminal
offense because of the creation or distribution of fake digital
media content that falsify realism~\cite{deepfakeact}.
Besides, social media platforms (\eg Twitter, Facebook) are actively taking initiatives to
deal with forged, synthetic, and manipulated content on their respective platforms.
For example, in Twitter, if a tweet contains manipulated media content specially, \name s, Twitter has started to alert users about that by tagging with warning sign and attaching the trustworthy news article link relevant to the tweet~\cite{twitterini}.
In another example, in 2019, Facebook facilitated the development of robust \name~detection tools by organizing the \name~detection challenge (DFDC) where 2114 number of participants across the globe had participated and they generated more than 35,000 models~\cite{dolhansky2019deepfake}.

\begin{figure} [h]
\centering
\includegraphics[width=.85\textwidth]{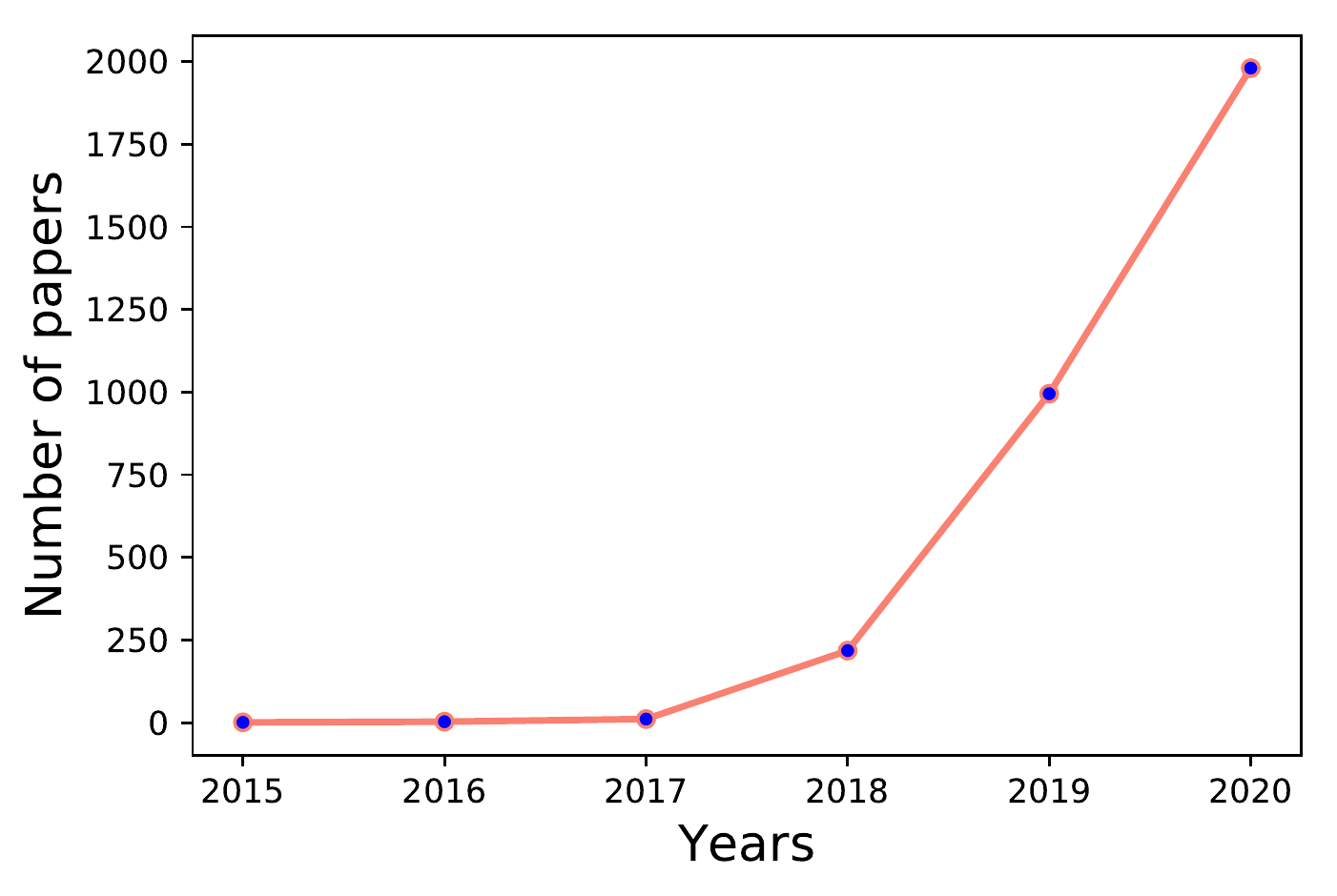}
\caption{Number of scholarly articles that are related to \name~in years from 2015 to 2020,
obtained from Google Scholar on April 2021 with the query
keyword “deepfake” applied to either title or full text of the papers.}
\label{fig:review}
\end{figure}

In Figure~\ref{fig:review}, we depict the relation between number of papers that are related to \name~in years from 2015 to 2020. The data is collected from Google Scholar on April 2021 with the query keyword “deepfake” found in either title or full text of the papers.
According to the number of related papers has increased
significantly in the recent years which is an indication Deep Fakes related research or news are getting noticed a lot more in recent times.

This chapter presents a handful number of methods for \name~generation and detection in comprehensive manner.
In Section~\ref{sec:deep generation}, we discuss how deep learning is leveraged on  
\name~algorithms for creating manipulated contents. 
Section~\ref{sec: deep detection} reviews
a wide set of effective methods for \name~detection as well as their
advantages and disadvantages.
In Section~\ref{sec:chal}, We discuss challenges, research trends, and directions on \name~generation and detection domains.
Finally, Section~\ref{sec:conclusion} concludes the study.

%% file: Sources/sec-generation.tex
\section{DeepFake Generation}
\label{sec:deep generation}
\name~contents get attraction because of the availability of robust and powerful set of \name~applications to a wide range of users. Such applications are capable to create a forged content by a few number of clicks within a few seconds. Because of their rapid popularity, a large number of researches have been performed on \name~generation in recent days.
In this section, we discuss about various \name~generation approaches and the datasets.

\subsection{Generation Approaches}
\label{subsec:deep gen app}
Most of the DeepFake related works have been done leveraging deep learning techniques. From the state-of-the-art literatures, \name~generation through facial image manipulation can be classified into four methodological categories based on the way and extent of manipulation: complete face synthesis, identity swap, attribute manipulation, and Face Reenactment. We provide a detailed discussion in the following sections. We discuss all other methods in a separate group - \lq\lq Other \name~Generation Methods\rq\rq. %Figure~\ref{fig:tax deep gen} depicts a taxonomy of the category-wise state-of-art \name~generation approaches. 

%PUT TAXONOMY FIGURE

\subsubsection{Complete Face Synthesis}

Facial manipulation or editing techniques have been studied and developed considerably over the last few decades. It is also practised for generating \name~contents in recent days. 
Face synthesis generates photorealistic images of human faces that do not exist in real life. 
With the massive progression in Generative models, GAN, \cite{goodfellow2014generative} in the last few years, the research community has seen a significant amount of works associated with facial manipulation. GANs have been effectively used for generating photorealistic face images. Another variation of the generative deep learning model is variational autoencoder (VAE) \cite{kingma2013auto} that also shows the potentiality in creating human face image.

Initially, these adversarial model starts generating realistic fake images from random vectors. The generative model tries to generate a more realistic image and fool the discriminative model in each iteration. On the contrary, the goal of the discriminative model is to verify the generated photo is either real or fake. 
Radford \etal proposed the deep convolutional generative adversarial network (DCGAN) \cite{radford2015unsupervised}, where the concept of both GAN and Convolutional Neural Network (CNN) has been utilized together to create a nonexisting human face.  It is one of the initial works after the emergence of GAN in 2014. Liu \etal proposed VAE based CoGAN \cite{liu2016coupled}. In COCO-GAN \cite{lin2019coco}, the authors proposed a conditional GAN-based image generator that is capable of synthesizing images in a parallelizable fashion. However, Glow \cite{kingma2018glow}, a flow-based generative model, which is different from GAN's mechanism, is proposed by Kingma e\etal. In this work, the authors used invertible 1x1 convolution for generating realistic \name~images.

 Later in 2017, Wasserstein generative adversarial networks (WGAN) \cite{arjovsky2017wasserstein} has been proposed. The approach used in WGAN training is more stable than the previous method. Stability in GANs training was one of the primary issues in the first few years right after GAN's invention. WGAN minimizes this instability in GAN training. However, Gulrajani \etal \cite{gulrajani2017improved} showed that due to the weight clipping operation, sometimes WGAN  might fail to converge, thus might generate lousy images as output. In this paper, they provide an improved weight clipping approach to address the issue in WGAN training. BEGAN \cite{berthelot2017began} is another work with the aim of improving WGAN. Karras \etal presented Progressive Growing GAN (PGGAN) \cite{karras2017progressive} in 2017 with the focus on generating high-quality images. This is one of the pioneering works on generating high-quality images. The same author proposed StyleGAN \cite{karras2019style} in 2019 that can automatically learn the high-level attribute representation such as identity, pose to control different properties in generated images. StyleGAN2 \cite{karras2020analyzing}, the extended version of the previous work, was presented in 2020.

\subsubsection{Identity Swap}
Identity swap is one of the most common face manipulation research techniques associated with DeepFakes. This approach includes replacing the human face in the target content (image or video) with another face in the source content. The traditional face swap process can be performed in three phases. First, the face is required to be detected in both source, and target content that can be done with face detection \cite{zhang2016joint}, or object detection model \cite{liu2016ssd,redmon2016you}. The research community has seen numerous defensive \cite{akcay2018using} and offensive \cite{hossen2020object} applications with object detection techniques. After face or facial attributes detection in source and target content, the eyes, nose, eyebrows, mouth is replaced and adjusted and blended in term of lighting and color to minimize the difference between source and target content. In the third step, the adjusted candidates are ranked by the calculated distance over the overlapped region. However, this traditional face swap approach has limitations in generating very realistic face images as it offers static and rigid replacement. Different DL-based approaches have become very effective in realistic face-swapping with the super-progress in the Deep learning (DL) domain.

\begin{figure} [H]
\centering
\includegraphics[width=.9\linewidth]{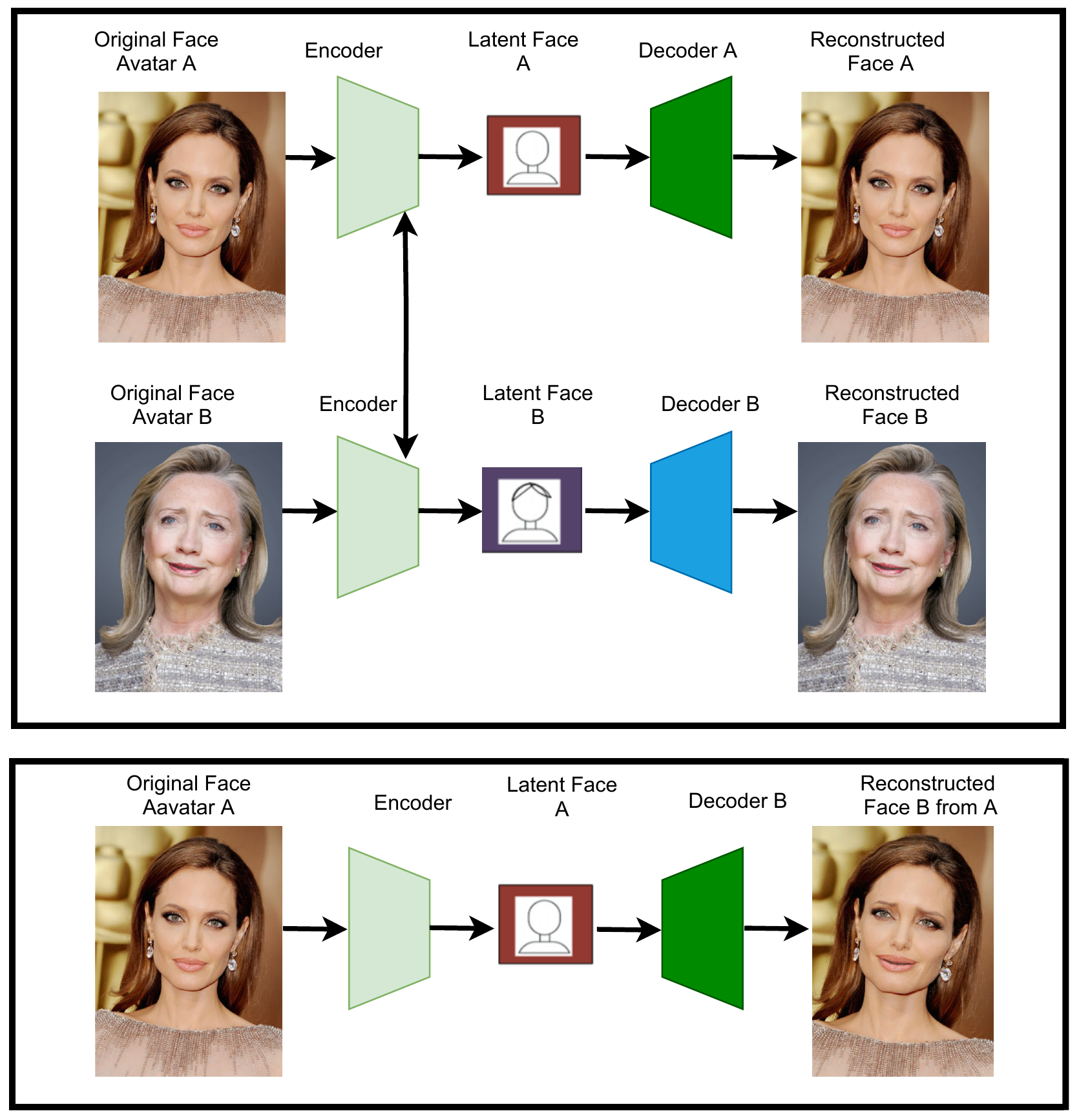}
\caption{A \name~creation model using two encoder-decoder combinations. In the training phases, the encoder-decoder set is used to learn the latent features of the input faces. While generation, two decoders are interchanged, such that latent face A is subjected to decoder B to generate the face A including the features of face B. The face images used in the figure are obtained from~\cite{masood2021deepfakes}.}
\label{fig:deep gen}
\end{figure}

FaceSwap \cite{FaceSwap}, and CycleGAN \cite{zhu2017unpaired} are some of the very first works in this field. In FaceSwap, two sets of encoder-decoder combinations are used. The encoder part of the architecture is responsible for composing the latent feature of a face from the input image, and then the decoder part reconstructs the face. There are two parts of the training phase. In the first phase, Each encoder-decoder combination is trained with the source image. In the second phase of the training, the decoder gets trained with the target image. After successful training, the two decoders are substituted with each other. Consequently, the original encoder paired with the decoder of the target image is capable of constructing the target image with the facial features of the source image. The DeepFake generation (identity swap) procedure with pairs of encoder-decoder architecture is illustrated in Figure \ref{fig:deep gen}.

The CycleGAN \cite{zhu2017unpaired} solved the issue of the unavailability of paired training examples for image translation. FSGAN \cite{nirkin2019fsgan} is capable of face swapping and reenactment simultaneously with face reenactment and blending. Natsume \etal proposed two distinct VAE based RSGAN \cite{natsume2018rsgan} to encode the latent representation of facial attributes. The recent work, FaceShifter \cite{li2019faceshifter} uses a two-phase scheme for high fidelity and occlusion-based face-swapping. 

\subsubsection{Attribute Manipulation}
Face attribute manipulation is a process of modifying the specific region of the face in the target image. This process is very similar to face editing in most cases. Some examples of face manipulation are changing age, gender, hair color and style, disappearing hair (bald), and creating smiling faces from neutral faces, etc. Choi \etal applied the GAN concept in the image-to-image translation problem by presenting StarGAN \cite{choi2018stargan} in 2018. In StarGAN, There is a single generator for translating images from a domain to multiple domains. However, StarGAN has a limitation as it can only generate a specific number of expressions. For Image translation, Chen \etal proposed HomointerpGAN \cite{chen2019homomorphic}, where during translation intermediate region between different domains is considered. The author suggested proper methods to select paths between two sample points in latent space to change particular image attributes.

Pumarola \etal mitigated this issue by proposing GANimation \cite{pumarola2018ganimation} where action unit (AU) annotation based GAN conditioning method is implemented. Later, the authors presented the improved version StarGANv2 \cite{choi2020stargan} that can generate images with the highest visual quality. To develope this improved version, the authors design the model with encoder-decoder architecture where a random Gaussian noise vector fed to the generator. In terms of expression synthesis and attribute manipulation, StarGANv2 outperforms other works for its high scalability.
To achieve better. He \etal introduced encoder-decoder-based AttGAN \cite{he2019attgan} where conditioned latent representation is used for more specific facial attribute editing as well as preserving other details of the face. One of the limitations of AttGAN is unwanted blurriness in generated face images. STGAN \cite{liu2019stgan} is proposed by Liu \etal in 2019 as an improvement of AttGAN. In this work, the difference between target and source attribute is taken into consideration for more specific face attribute editing. However, STGAN shows poor performance in multiple attribute manipulation in the face image. 

\subsubsection{Face Reenactment}
Face reenactment is a type of emerging \name~face manipulation technique, more precisely, which can be stated as a conditional face synthesis task for facial expression transfer. It refers to a process that replaces or transfers the facial expression of a person to another person. Face reenactment can be achieved by transferring the source actor's expression, gaze, pose, and mouth movement. Some of the most prominent works for real-time facial expression transfer have been done by Thies~\etal~\cite{Thies2019Face2FaceRF,roessler2019faceforensicspp}. In Face2Face, the authors proposed methodologies to transfer the source person's facial expressions (actor), including facial gestures, head, and eye movement, to a video with another person (target person), maintaining the identity. Face2Face approach use deformation transfer between source face and target face, more specifically the mouth portion with higher priority for photo-realistic reenactment. For the tracking and reconstruction of the face identity (3D face model) of a source and target model, a commodity RGB-D sensor is used. After getting the required parameters from the 3D models, face expression is reenacted to the target face in each generated fake video frame. This approach is applicable in real-time standard RGB videos. In the same research group's successive work, FaceForensic++~\cite{roessler2019faceforensicspp} has been presented where NeuralTextures based learning approach has been utilized. 

In the last few years, the research advancement in the development of GANs (Generative Adversarial Network) \cite{goodfellow2014generative} is remarkable. GANs are comprised of two models, a generative model and a discriminative model, that are estimated through the adversarial process. Both models compete with each other to minimize their loss function in a fashion that can be interpreted as a minimax two-player game. GANs have been proven to be very effective for facial reenactment to generate realistic examples across a wide range of domains. The authors employ conditional GANs (cGANs) as a solution to image-to-image translation problems in their work \cite{isola2018imagetoimage} that is identical to the techniques from pix2pix software. Another work, Pix2pixHD \cite{wang2018highresolution}, where authors proposed a successful high-resolution image generation approach using multi-scale cGANs with a perpetual loss. Wu \etal proposed ReenactGAN \cite{Wu2018ReenactGANLT} to transfer both mouth and expression to the target face by mapping the source face boundary latent to the target face's boundary latent with a transformer. Eventually, reenacted target face is generated in a fake video with a target-specific decoder. 

Zhang \etal \cite{Zhang2019OneshotFR} proposed a one-shot approach to generated reenacted faces using only a single source image. The authors presented an auto-encoder-based model that can learn a latent representation of both the source and target face representation. A similar but recent one-shot face reenactment model, FaR-GAN \cite{Hao2020FaRGANFO} has been proposed by Hao \etal.

\subsubsection{Other Generation Methods}
Other than previously discussed approaches, some works with different methodologies might be classified as \name~generation approaches. This category includes lip-syncing, inpainting, style transfer, super-resolution, etc. 

Lip-syncing \name~video generation approach produces a video of a target person in a fashion that the mouth and lip movement in the output video is synchronized with a source audio input. This synchronized mouth region movement makes the generated video realistic. Fan \etal introduced a deep bidirectional Long short-term memory (LSTM) based approach \cite{fan2015photo} for audio/visual modeling to develop a photo-real talking head system. LSTM is a subset of recurrent neural network (RNN) architecture that can model sequential data where long-term dependencies need to be considered.  LSTM is widely adopted in prediction from health data \cite{rabby2021stacked}, natural language processing (NLP) \cite{zobaed2021senspick}, next video frame prediction \cite{hosseini2020inception} etc. LSTM model with other DL models can learn and predict the lip movement from the input source audio file. 

The image inpainting approach involves reconstructing the missing or incomplete part of images or videos. Yu \etal presented a well-known work, ContextAtten \cite{Yu_2018_CVPR} in 2018 in image inpainting. The most common issue in the previous image inpainting is distorted structures or blurry texture in the manipulated image. Later, this issue is addressed in ContextAtten. SC-FEGAN \cite{jo2019sc} by Jo \etal is an image editing work focusing on utilizing a relatively free-form user input in terms of color and shape.

Single image super-resolution (SISR) task might be considered as a variation of \name~generation. Dai \etal proposed the Second-order attention network (SAN) \cite{dai2019second} for more effective feature expression and correlation learning. In this work, the authors focus on feature correlation instead of using a model with deep architecture. Karnewar \etal proposed MSGGAN \cite{karnewar2020msg} for high-resolution image synthesis with a goal of achieving well convergence on a variety of image datasets. For increasing convergence stability, The authors allow the gradients-flow from the discriminator to the generator at various scales. 

One of the most common recognizable factors in \name~images is artifacts in their frequency domain. Some of the DeepFake research community try to eliminate traceable artifacts by modifying the generation procedure. SDGAN \cite{jung2020spectral}, WUCGAN \cite{durall2020watch} are examples of such works. In WUCGAN, a spectral regularization has been used to overcome the GANs' inability to produce real image spectral distribution due to the up-sampling method.

In Table~\ref{table:generation-approaches}, we provide summaries of the \name~generation works mentioned above.

\begin{table}[H]
\caption{Summary of DeepFake Generation approaches. We list 26 recent approaches published in peer-reviewed journals and conferences in the table.}
\label{table:generation-approaches}
\resizebox{\linewidth}{!}{
\begin{tabular}{|c|c|c|c|c|c|}
\hline
\multirow{2}{*}{\textbf{Works}}                                                              & \multirow{2}{*}{\textbf{Methods}} & \multirow{2}{*}{\textbf{Elo Rating}} & \multirow{2}{*}{\textbf{Datasets}} & \multicolumn{2}{c|}{\textbf{Multimedia}}    \\ \cline{5-6} 
                                                                                     &                         &                             &                           & \textbf{Image} & \multicolumn{1}{l|}{\textbf{Video}} \\ \hline
\cite{kingma2018glow} & Glow                    & 1511                        & \small{CIFAR-10, ImageNet, LSUN}                         & \cmark     &  \xmark                         \\ \hline
\cite{jo2019sc}               & SC-FEGAN                & 1489                        & \small{CelebA-HQ}                      & \cmark      & \xmark                          \\ \hline
\cite{radford2015unsupervised}     & SAN                     & 1487                        & \small{LSUN, Imagenet, Faces, CIFAR-10, SVHN, MNIST}                   &  \cmark     &  \xmark                          \\ \hline
\cite{gulrajani2017improved}    & WGAN-GP                 & 1435                        & \small{LSUN, Google
Billion Word, Swiss Roll}                       & \cmark     & \xmark                           \\ \hline
\cite{yu2018generative}                & ContextAtten            & 1430                        & \small{CelebA, CelebA-HQ, ImageNet, Places2, DTD}                         & \cmark      & \xmark                           \\ \hline
\cite{he2019attgan}                    & AttGAN                  & 1426                        & \small{CelebA, LFW}                         &  \cmark    & \xmark                      \\ \hline
\cite{lin2019coco}                    & CocoGan                 & 1400                        & \small{CelebA, CelebA-HQ, LSUN,  Matterport3D}                         & \cmark      & \xmark                            \\ \hline
\cite{durall2020watch}             & WUCGAN                  & 1400                        & \small{FaceForensics++, CelebA, Faces, faces-HQ}                         & \cmark     & \cmark                            \\ \hline
\cite{zhu2017unpaired}                & CycleGAN                & 1400                        & \small{Cityscapes,  CMP Facade, UT Zappos}                         & \cmark      & \xmark                           \\ \hline
\cite{pumarola2018ganimation}    & GANimation              & 1390                        &  \small{EmotioNet, RaFD}                         & \cmark      & \xmark                          \\ \hline
\cite{choi2018stargan}               & StarGAN                 & 1387                        &  \small{CelebA, RaFD}                         & \cmark      & \xmark                           \\ \hline
\cite{karnewar2020msg}           & MSGGAN                  & 1385                        &  \small{CelebA-HQ, CIFAR-10, OF, LSUN, FFHQ}                         & \cmark      & \xmark                          \\ \hline
% \cite{park2019semantic}              & GauGAN                  & 1383                        & \small{COCO, ADE20K, Cityscapes, Flickr}                          & \cmark       & \xmark                           \\ \hline
\cite{mao2016multi}                   & DCGAN                   & 1337                        &  \small{HWDB1.0, LSUN, MNIST}                          & \cmark     & \xmark                           \\ \hline
\cite{chen2019homomorphic}           & HomointerpGAN           & 1335                        & \small{RaFD, CelebA}                         & \cmark      &  \xmark     
\\ \hline
\cite{karras2017progressive}           & PGGAN           & 1336                        & \small{CelebA, LSUN, CIFAR10}                         & \cmark      &  \xmark
\\ \hline

\cite{berthelot2017began}           & BEGAN           & 1291                        & \small{CelebA}                         & \cmark      &  \xmark
\\ \hline

\cite{natsume2018rsgan}           &    RSGAN        &  N/A                       & \small{CelebA}                         & \cmark      &  \xmark
\\ \hline

\cite{Hao2020FaRGANFO}           &    FaR-GAN        &  N/A                       & \small{VoxCeleb1}                         & \cmark      &  \xmark
\\ \hline

\cite{jung2020spectral}           &    SDGAN        &  N/A                       & \small{FFHQ}                         & \cmark      &  \xmark
\\ \hline

\cite{liu2019stgan}           &    STGAN        &  N/A                       & \small{FFHQ}                         & \cmark      &  \xmark
\\ \hline

\cite{Wu2018ReenactGANLT}           &    ReenactGAN        &  1400                       & \small{CelebV, WFLW}                         & \xmark      &  \cmark
\\ \hline

\cite{choi2020stargan}           &    StarGANV2        &  1400                     & \small{CelebAHQ, AFHQ}                         & \cmark     & \xmark
\\ \hline

\cite{karras2019style}           &    StyleGAN        &  N/A                       & \small{FFHQ}                         & \cmark      &  \xmark
\\ \hline

\cite{karras2020analyzing}           &    StyleGAN2        &  1416                       & \small{FFHQ}                         & \cmark      &  \xmark
\\ \hline

\cite{nirkin2019fsgan}           &    FSGAN        &  1400                       & \small{IJB-C, VGGFace2, Figaro, Forensics++, CelebA }                         & \cmark      &  \cmark
\\ \hline

\cite{liu2016coupled}           &    CoGAN        &  N/A                       & \small{MNIST, USPS, CelebA, RGBD, NYU}                         & \cmark      &  \xmark
\\ \hline

\end{tabular}
}
\end{table}

\subsection{DeepFake Generation Dataset}

A proper dataset plays a pivotal role in the deep learning model's performance. For DeepFake generation, two major types of datasets are used for different purposes: real dataset and synthesized dataset. In most cases, real datasets are used for DeepFake generation, whereas synthesized or fake datasets are used for DeepFake detection. Discussion on popular real datasets in the provided section below.

Yi \etal presented CASIA-WebFace \cite{yi2014learning} in 2014. The dataset includes 10,575 subjects and 494,414 images. CelebA \cite{liu2015deep} dataset was introduced by Liu \etal. This is a labeled version of CelebFaces \cite{sun2013hybrid}. In CelebA dataset, there are 10,000 subjects, a total of 200,000 images where each subject has twenty samples.  

VGGFace \cite{parkhi2015deep} is a large dataset with 2.6 million images of 2,622 subjects. Microsoft Celeb (MS-Celeb-1M) \cite{guo2016ms} is another large image dataset, particularly for face recognition. This dataset contains 10 million face images of 100k identities. Cao \etal introduced VGGFace2 \cite{cao2018vggface2} in 2018, containing 3.31 million images of 9131 persons. Images are collected from the internet. This dataset has a wide variation in age, ethnicity, and pose. Flickr-Faces-HQ \cite{karras2019style} was presented in 2019 by Karras \etal containing 70k high-resolution images.

%% file: Sources/sec-detection.tex
\section{DeepFake Detection}
\label{sec: deep detection}
It is obvious that \name s can be tremendous threats ranging from a person to the whole world. To avoid the threats, an effective set of \name~detection approaches
are required. Subsequently, an increase amount of research is performed for developing various approaches to detect the authenticity of a still image or video content.
In the past, a manipulated content was determined manually by analyzing artifacts and inconsistencies. In recent days, because of leveraging deep learning, complex and discriminative features are extracted to detect fake contents.

%If I have time, I will provide taxonomy. Otherwise, will delete taxonomy from creation.

\subsection{Detection Approaches}
In this section, we review recent studies on various \name~detection works, based on their methodologies and extracted attributes.

\subsubsection{Forensics-based Detection}
Recent forensics-based detection studies analyze pixel-level disparity.
In addition, they provide explainable
detection mechanism to determine authenticity. 
however, these works undergo robustness issues when the manipulated contents are generated by simple transformations.   

Li \etal observed that the disparities between manipulated and real faces are revealed in the color components~\cite{li2020identification}. 
The authors proposed training a one-class classifier on real face data based on considering the
disparities in the color components to detect the unknown GANs. 
However, they did not clarify the performance of their approach against perturbation attacks such as image transformations.

In~\cite{koopman2018detection}, Koopman \etal proposed to detect fake videos based on the unique noise pattern in the videos that is caused by the camera sensor. 
Rather than considering noise, the authors of~\cite{yang2019exposing} observed
that \name s usually contain inconsistent or unusual head poses with respect to expression.
Hence, their work monitors facial landmarks to calculate the differences in head pose between manipulated and genuine video frames.
They use the difference in
estimated head pose data as a feature vector to train an SVM
based classifier to predict original and \name s.
Unfortunately, the authors did not clarify the effectiveness of their works~\cite{koopman2018detection,yang2019exposing} in detecting high quality \name s.

In~\cite{wang2020cnn}, Wang \etal leveraged
the local motion features captured from real videos to identify
the inconsistency of forged videos.
They emphasized on the low-level
features that are not feasible to be deployed in the wild
where \name s suffer known and unknown degradations.

Demir \etal \cite{demir2021deep} focused on synthetic eyes construction in deep fake videos. %Using eye and gaze features they achieved an average of 86 \% accuracy on different benchmark datasets(FaceForensics++, DeepFakes and Celeb-DF).
%I added from here to the down.
They generated features from eye and gaze data to train their model and compare it with complex state-of-the-art CNNs ( VGG19, Inception, Xception, ResNet, and DenseNet).
They claimed fake detection accuracy is around $6.5\%$ higher than those of complex architectures without using eye or
gaze information. However, considering only gaze data for detecting a synthesized content does not indicate improvement in generalization. Therefore, it is not clear that how the model will perform in detecting unseen adversary.  

%Along side with DeepFake video generation, audio spoofing is another wayof Character assassination of a public figure. Chintha et  al.[90] addressed this problem  and  proposed  a  convolutional  bidirectional  recurrent  architecture  for both DeepFake video detector and audio spoof detector

\subsubsection{Deep Neural Network-(DNN) based Detection}
For DeepFakes detection in images and videos, neural network models with deep architecture outperform classical/hand-crafted approaches. DNN based models are capable of learning meaningful features from available data for effective forgery detection. Guera \etal \cite{guera2018deepfake} proposed a \name~video detection framework model combined with an RNN and CNN architecture to detect forged part from the input video. 
However, the limitation of the work is its incapability to handle videos for more than 2 seconds. 

Nataraj \etal \cite{nataraj2019detecting} presented a CCN model-based approach to calculate pixel co-occurrence matrices from the input image to detect image manipulation. Nguyen \etal \cite{nguyen2019multi} proposed a robust \name~video detector with multi-task CNN base architecture to identify and localize the manipulated portions from a video. An autoencoder and a decoder are used for the classification of manipulated content and sharing the extracted features for segmentation and shape reconstruction, respectively. However, the accuracy of this model declines for unseen examples that can be considered as a limitation of the work. To address this accuracy degradation-related limitation, a Forensic Transfer (FT) based CNN approach
\cite{cozzolino2018forensictransfer} for \name~detection was proposed by Stehouwer \etal. Marra \etal \cite{marra2019incremental} presented an approach based on incremental learning for GAN-generated fake image detection. This work focuses on the classification and detection of a new type of GAN-generated images with high accuracy. This approach is more generalized and robust against detecting unseen GAN-generated examples. However, This procedure still needs some information about the new GAN architecture, which affects the practicality of the work.

% \subsubsection{Preprocessing-based Detection}
% % \textbf{Facial image preprocessing} \\
% DeepFake detection becomes more challenging when in a video frame multiple faces are observed. Charitidis \etal \cite{symeoninvestigating} tried to solve this problem with the improvement of preprocessing step. According to their approach they prune cluster of facial data which carries less significance. This approach makes DeepFake detection process fast and it can be used on top of any kind of existing DeepFake detection system. This approach still have one problem, their preprocessing approach can discard less prevalent but significant data from the datasets. 

\begin{table}[]
\centering
\caption{We present Summary of existing top-25 notable \name~detection work published in peer-reviewed journals and conferences in the table. We report only the highest achieved resultant metrics. ACC, PRE, AUC, and EER denotes accuracy, precision, area under the curve, and equal error rate, respectively.}
\label{table:detection-approaches}
\resizebox{\linewidth}{!}{
\begin{tabular}{|c|c|c|c|c|c|}
\hline
\multicolumn{1}{|c|}{\multirow{2}{*}{\textbf{Works}}} &
  \multicolumn{1}{c|}{\multirow{2}{*}{\textbf{Methods}}} &
  \multicolumn{1}{c|}{\multirow{2}{*}{\textbf{Performance}}} &
  \multicolumn{1}{c|}{\multirow{2}{*}{\textbf{Datasets}}} &
  \multicolumn{2}{c|}{\textbf{Multimedia}} \\ \cline{5-6} 
\multicolumn{1}{|c|}{} &
  \multicolumn{1}{c|}{} &
  \multicolumn{1}{c|}{} &
  \multicolumn{1}{c|}{} &
  \multicolumn{1}{c|}{\textbf{Image}} &
  \multicolumn{1}{c|}{\textbf{Video}} \\ \hline
  \cite{li2020identification}            & One-class   & ACC: 0.98  & Self-built                  & \cmark & \xmark \\ \hline
\cite{khalid2020oc}           & VAE   & ACC: 0.98 & FF++                  & \xmark          & \cmark \\ \hline
\cite{ciftci2020fakecatcher}  & CNN   & ACC: 0.96 & FF, FF++, Celeb-DF    & \xmark          & \cmark \\ \hline
\cite{li2020sharp}            & S-MIL & ACC: 0.83 & FF++, Celeb-DF, DFDC  & \xmark          & \cmark \\ \hline
\cite{masi2020two}            & RNN   & AUC: 0.99 & FF++, Celeb-DF, DFDC  & \xmark          & \cmark \\ \hline
\cite{feng2020deep}           & CNN   & AUC: 0.99 & UADFV, Celeb-DF, FF++ & \xmark          & \cmark \\ \hline
\cite{dang2020detection}      & CNN   & ACC: 0.98 & Celeb-DF, UADFV, DFFD & \cmark & \xmark          \\ \hline
\cite{mittal2020emotions}     & DNN   & AUC: 0.96 & TIMIT, DFDC           & \xmark          & \cmark \\ \hline
\cite{chai2020makes}          & CNN   & PRE: 1.0  & FF++                  & \cmark & \cmark \\ \hline
\cite{li2020face}             & HRnet & ACC: 0.95 & UADFV, Celeb-DF, FF++ & \xmark          & \cmark \\ \hline
\cite{tarasiou2020extracting} & CNN   & ACC: 0.98 & FF++                  & \xmark          & \cmark \\ \hline
\cite{nguyen2019multi}        & CNN   & ACC: 0.93 & FF++                  & \xmark          & \cmark \\ \hline
\cite{afchar2018mesonet}      & CNN   & ACC: 0.98 & FF++                  & \xmark          & \cmark \\ \hline
\cite{sabir2019recurrent}     & RNN   & ACC: 0.99 & FF++                  & \xmark          & \cmark \\ \hline
\cite{wang2020cnn}            & CNN   & PRE: 1.0  & FF++                  & \cmark & \cmark \\ \hline
\cite{guera2018deepfake}            & RNN   & ACC: 0.97  & self-built                  & \xmark & \cmark \\ \hline
\cite{koopman2018detection}            & N/A   & N/A  & self-built                  & \xmark & \cmark \\ \hline
\cite{nataraj2019detecting}            & CNN   & ACC: 0.99  & self-built                  & \cmark & \xmark \\ \hline
\cite{demir2021deep}            & CNN   & ACC: 0.89  &   FF++                & \xmark & \cmark \\ \hline
\cite{chintha2020recurrent}            & CNN   & EER 0.13  &   FF++, Celeb-DF                 & \xmark & \cmark \\ \hline
\cite{cozzolino2018forensictransfer}            & CNN   & ACC 1.0  &   Self-built                 & \cmark & \xmark \\ \hline
\cite{zhang2019detecting}            &CNN    & ACC: 1.0   & Self-built                    & \cmark & \xmark \\ \hline

\cite{mccloskey2019detecting}  & SVM   & AUC: 0.91   &              Self-built      & \cmark & \xmark \\ \hline

\cite{marra2019incremental}  & N/A    & N/A   &     Self-built      & \cmark & \xmark \\ \hline
\cite{yu2019attributing}  & CNN    & ACC: 0.99   &     Self-built      & \cmark & \xmark \\ \hline

\end{tabular}
}
\end{table}

\subsubsection{GAN Redesign-based Detection}
In lieu of considering only audio-visual artifacts,
a few number of research criticize the design limitation of existing GAN-based approaches and highlight to redesign GAN by including new artifacts. 

In~\cite{mccloskey2019detecting}, McCloskey and Albright investigated the traditional architecture
of the generator function and observed that the internal values
of the generator are normalized. They claimed the normalization technique limits the frequency of
the saturated pixels and makes it difficult to calculate the occurrences of saturated and underexposed pixels.
They suggested to use their proposed approach as a complementary to other approaches that
detect visual artifacts in the manipulated contents.

Zhang \etal investigated how the generalization ability of the existing detectors are impacted due to the existing upsampling design related artifacts~\cite{zhang2019detecting}. They also noted that upsampling design is generic in GAN pipelines. Hence, they proposed a new signal processing analysis and redesigned the classifier accordingly. 
In addition, they developed a simulator
framework, AutoGAN that simulates the common generation pipeline
shared by a large class of popular GAN models~\cite{zhang2019detecting}. AutoGAN
simulates the GAN generation pipeline and generates (simulated) fake images that can be used in training any classifier without the burden of accessing pre-trained GANs.

In~\cite{yu2019attributing}, Yu \etal proposed GAN fingerprint artifact for classifying the images and also determining the source of a target images. Although an insignificant amount of differences yield a distinct fingerprint, the fingerprints is vulnerable (\ie tempering) to perturbation attacks such as image transformation, blur, JPEG compression, and so on.

\subsubsection{Visual and Audio Inconsistency-based Detection}

Mittal \etal \cite{mittal2020emotions} work has addressed the essentiality of multimodal approach for DeepFake detection. They proposed an approach combining two modalities: the audio(speech) and video (face) to extract emotional features from both modalities to detect any kinds of counterfeit in the input video. This approach won’t work if multiple persons are present in one video. 

In~\cite{zhang2021attribution}, the authors  tried to buck forgery in the realm of videos and images, by inspecting/traceback the source/mechanism of a given DeepFake image. ML tools are not always enough to combat this kind of problem. In addition, current robust DeepFake detection systems are vulnerable to adversarial images/videos.  For these reasons, instead of building a robust DeepFake image/video detection system, it is more effective and scalable to find the associated generative model. With the help of a trusted third party, we can restrict/limit the malicious purpose of usage of this model. But deniability, misattribution to the original developer still a problem of the attribution approach. In easy terms, we can mention this attribution process as \emph{Traitor Tracing}. This system will enforce accountability among model developers.

Along side with DeepFake video generation, audio spoofing is another way of character assassination of a public figure. Chintha \etal \cite{chintha2020recurrent} addressed this problem by finding inconsistencies in audio and video modalities. To this end, the authors leveraged XceptionNet  architecture for facial feature extraction and stacked convolutional layers to generate audio embedding features. Our analysis suggests that the combination of spoof audio and fake video detection is prone to achieve better generalization that indicates robustness in detecting unknown adversaries.

\subsubsection{Other Notable Detectors}

Rashmiranjan \etal (2021) in~\cite{detecting2021euler}, investigated a technique involving Euler video magnification (EVM) process extracting features using three techniques (SSIM, LSTM, Heart Rate Estimation) to train models to classify counterfeit and unaltered videos. 
This technique uses spatial decomposition and temporal filtering on video data to highlight and magnify hidden features such as pulse of skin or subtle motions. 

Fernandes \etal applied similar technique in~\cite{fernandes2019predicting}, where they used EVM in color-based photoplethy smography (PPG) to identify blood volume fluctuations by shining light of certain wavelength onto the skin and measuring changes in light assimilation of the oxygenated blood which is in turn measures the heart rate. 
On the other hand, in this work, the authors applied both the EVM based color and movement amplification on videos to distinguish between original and fake videos. The results using SSIM technique when used a range of standard machine learning shows below 82\% accuracy achieved by the best performing submission to the DFDC while the results using LSTM technique establishes apparent setback to the idea of using EVM for \name~detection. Overall, even though the color and spatial aspects of EVM were tested as possibilities for a number of classification models, the authors used accuracy as a metric though it is not known to be great metric for evaluation when imbalanced datasets are used which can be improved.

Hussain \etal \cite{hussain2021adversarial} and Carlini \etal \cite{carlini2020evading} discussed the vulnerability of current DeepFake detectors in light of both the whitebox and blackbox attacking approach. Carlini \etal \cite{carlini2020evading} also demonstrated that a novice attacker can effectively conduct a blackbox attack without having any information regarding classifier and can reduce classifier’s AUC to 0.22.

DeepTag~\cite{wang2020deeptag} is another digital watermarking-based proactive system to combat DeepFake problems. This system finds the source of a DeepFaked image with an embedded message associated with the original image. This system works better against the dynamic image transformation and reconstruction of images by the DeepFake process. By blocking the confirmed DeepFake, this system also helps to stop spreading misinformation on the different social media platforms. According to their approach, the embedded message has to avoid the manipulated region. Even though the authors addressed this problem, they did not provide any solution on this. 

DeepFake detection becomes more challenging when multiple faces are observed in a video frame. Charitidis \etal \cite{symeoninvestigating} tried to solve this problem with the improvement of preprocessing step. They pruned a cluster of facial data that carries less significance. This approach makes DeepFake detection process fast and it can be used on top of any existing DeepFake detection system. This approach still has one problem, their preprocessing approach can discard less prevalent but significant data from the datasets. 

Table~\ref{table:detection-approaches} shows the summary of the aforementioned \name~detection works and corresponding dataset information.

\subsection{DeepFake Detection Dataset}
\textbf{The DeepFake Detection Challenge (DFDC) Preview Dataset}
\newline
In~\cite{dolhansky2019deepfake}, Dolhansky~\etal introduced a preview of DFDC dataset containing 5,000 videos that featured two facial modification algorithms where the actors were in agreement to use and manipulate their likeliness. To ensure visual variability, diversity in several axes (gender, skin tone, age) and arbitrary background was considered. A reference performance baseline was provided in terms of specific metrics that was defined and tested on two existing models for detecting \name s. The initial baseline consists of the performance check of three simple detection models. The first model is trained to detect low-level image and the other two models were trained on the FaceForensics++ dataset~\cite{rossler2019faceforensics++} and evaluated as implemented in~\cite{roessler2019faceforensicspp}. All performances of these three sample detection models were analyzed using precision, recall, and the logarithmic scale of weighted precision to detect half, most, or nearly-all \name s. 

\textbf{The Celeb-DF Dataset}
\newline
At least until the year 2019, \name~datasets included low visual quality and had little to no resemblance to \name~videos found online. The work presented in~\cite{li2020celeb} has constructed a large scale \name~video dataset called \textit{Celeb-DF} that includes a total of 5,639 high-quality \name~videos, corresponding to more that 2 million frames from publicly available YouTube video clips of 59 celebrities of diverse genders, ages, and ethnic groups using improved synthesis process. The video quality in \textit{Celeb-DF} with very few notable visual artifacts have significant differences with then available \name\, videos available online that included low-quality synthesized faces, visible splicing boundaries, color mismatch, and inconsistencies in synthesized face orientation etc. The overall quality of the videos were enhanced in terms of improving low resolution of synthesized faces, color mismatch, inaccurate face masks, and temporal flickering. The authors also presented a comprehensive evaluation with 9 \name\, detection methods and datasets considered making the most comprehensive study of \name\, detection available by then. Overall, this \textit{Celeb-DF} dataset has helped lowered the gap in the video quality between the actual and \name\, datasets that can be found online with a possibility of enlarging \textit{Celeb-DF} and further enhancing the visual quality including the running efficiency. 

\textbf{The FaceForensics++ dataset}
\newline
Rossler \etal in~\cite{rossler2019faceforensics++} generated a large-scale dataset with an automated benchmark based on classical computer-graphics and learning-based based methods such as \name s~\cite{Deepfakes_git}, FaceSwap~\cite{FaceSwap_git}, NeuralTextures~\cite{thies2019deferred}, and Face2Face~\cite{thies2016face2face}. This benchmark contains a hidden test set of an order magnitude larger than comparable, publicly available dataset including 1.8 million manipulated images extracted from 1,000 real videos and target ground truth to enable supervised learning. The authors conducted a thorough study of data-driven forgery detectors and showed that the use of domain specific information in conjunction with a XceptionNet classifier improves the detection with an unprecedented accuracy. This work also presented ways to automatically detect any forms of facial identity and facial expression manipulations with an automated benchmark consisting with random compression and dimensions.

\textbf{The UADFV dataset}
\newline
Up until the revelation of the popular software \lq \lq DeepFake\rq\rq that used generative adversary networks (GANs), since any form of manipulation of videos/images involved huge time consumption for editing operations, realistic high quality fake videos were not widespread. Due to this software, ample of high-quality fake video flooded the internet and thus detecting such videos became important. In~\cite{li2018ictu}, the authors used 50 YouTube videos that lasted 30 seconds each representing one individual with at least one blinking occurred, to form the Eye Blinking Video (EBV) dataset. The left and right states of each frame were annotated with a user-friendly annotation tool. The training dataset that was used in this work is CEW~\cite{song2014eyes} which includes 1,793 images of closed eyes and 1,232 images of open eyes to train front-end CNN model, 40 videos as training set for the overall Long-term Recurrent Convolutional Networks (LRCN) model \cite{donahue2015long} and 10 videos as the testing set. This LRCN method was shown to exhibit best performance 0.99 compared to other methods such as Eye Aspect Ratio (EAR)~\cite{cech2016real} with performance 0.79 and CNN with 0.98. 

%% file: Sources/sec-future.tex
\section{Challenges and Opportunities for Future Research Direction}
\label{sec:chal}
We review over 60 articles published either in peer-reviewed journals and conferences or posted on arXiv regarding \name~generation and detection. In this section, we describe our observations including challenges, limitations, and new research scopes, after reviewing the studied articles.     
This will contribute on the future research in creating more realistic and
detection-evasive \name, and also sophisticated \name~detection model.

\subsection{Findings in \name~Generation}
\begin{itemize}
    \item  We notice low resolution and poor quality in the output image irrespective to any existing \name~generation work. Currently,
    generation of high-resolution and sharp images is difficult since such image makes the job easier to differentiate it from training images~\cite{odena2017conditional}.
    A deeper analysis claims that this causes spike in
the gradient problem and affects training stability~\cite{karras2018progressive}. 
To mitigate the challenge, PGGAN is proposed to grow generator and discriminator progressively. It starts from low-resolution images
and gradually, adds new
layers for higher-resolution (\ie details) as the training progresses~\cite{karras2018progressive}. However, PGGAN is still in premature stage and capable of generating only $(1024 \times 1024)$ size images.  

\item 
The attribute manipulation methods are limited as these can only change the properties followed by the training set.
 Therefore, such an attribute manipulation method is needed that could capture attributes are independent attributes to the training set.

\item In most of the cases, identity swap and expression swap do not consider the continuity of the videos. They neither consider gesture nor physiological signals such as eye blink, breathing frequency, heart beat, and so on. 

\item The fake datasets are expanded only considering the diversity of the content-related factors such as age, gender, background, and so on. Our investigation conform factors such as added noise (quality degradation), Gaussian blue, JPEG compression, contrast change, and so on can enhance the diversity in the dataset. DeeperForensics-1.0~\cite{jiang2020deeperforensics}  dataset offers image-level degradations but it is added artificially by post-processing. We expect natural image/video-level degradations (\ie over/underexposed photos, bit-rate variations, choices of codec) in the future generation of the dataset.

\end{itemize}

\subsection{Findings in \name~Detection}
\begin{itemize}
    \item  Most of the existing works generate image dataset to evaluate the effectiveness of their approaches leveraging various GANs. A large portion of the works do not unveil the details about the datasets that used in evaluation. Hence, the quality of the generated forged images still remains unknown.
    On the contrary, these works claim their competitive results in detecting various synthesized images built on their own. We emphasize on the development of public GAN-synthesized fake image dataset.
    
    \item For the sake of performance evaluation, existing works implement simple baselines (\eg  vanilla DNN-based methods) and compare it with their works rather than considering state-of-the-art ones. Claiming the superiority of their works by comparing with na\"ive approaches indicate biased evaluation. We expect that future works should be comparable to state-of-the-art works so that we can understand effectiveness of the proposed work.

\item  The aim of DeepFake detection research 
is to develop more robust and generalized
approaches. Subsequently, the research community is trying their level best to come up with effective approaches.
However, the recent works are simply evaluated
on simple DeepFake video datasets, such as FaceForensics++ (FF++).
We emphasize that future works should focus more on challenging
datasets for acceptable performance evaluation.  

\item Almost all of the existing studies report their experimental results by merely considering
the detection accuracy without reporting other popular metrics such as precision, recall, and the
relation with the quality of DeepFakes. To conduct an acceptable performance evaluation, a comprehensive experimental result set is mandatory. A robust set of experiments should contain the result of various effective metrics. Currently, there does not exist any metric that can measure DeepFake quality. 
We hope, in future, the researchers would come up with a new
metric for measuring the quality of DeepFakes.

\item Detecting the emerging unknown DeepFakes is crucial in today's world. Hence, developing a practical DeepFake
detector that is deployable in the wild is a necessity.
Towards developing an effective \name~detector, we observe a set of key factors that are: (1) advance generalization capabilities, robust against various attacks(\eg iamge/video transformations, adversarial attacks), and presenting explainable \name~detection result. 
Unfortunately, in reviewing
the recent DeepFake detection articles, we
find that the researchers simply ignore to evaluate the capabilities
of their works from the aforementioned perspectives.
    
\end{itemize}

%% file: Sources/sec-concl.tex
\section{Conclusion}
\label{sec:conclusion}
Due to great progress on generative deep learning algorithm in recent years, nowadays it has become a real challenge to identify the authenticity of any visual content found online~\cite{mirsky2021creation}.
The aim of creating the synthesized contents is either for malicious intent or just for fun. 
%any such activities are proven to spread misinformation creating a mockery of political leaders or celebrities that may cause serious instability in social, political, or in the target person’s everyday life and become a reason of defamation. 
To resist any unexpected scenarios such as creating a manipulated content of important persons (\eg political leaders, celebrities) or generate synthesized contents for a useful purpose, the current \name~research community needs to consider the existing published articles both in \name~generation and detection to plan for extensive research efforts in the future. 
In light on this and to make our understanding better, in this current work, 
our investigation shows that in recent years deep learning research community have been trying to solve two large research domains including \name~detection and generation related to \name~image and video contents. 
We have shed light on these domains by discussing state-of-the-art research works. We also try to depict how the research community shifts their attention from feature-based DeepFake detection to feature agnostic and policy-based approaches to combat evasion of DeepFake detection. We also provide comprehensive descriptions of different prominent datasets to facilitate researchers to determine their next research direction.  